\documentclass[10pt,journal,compsoc]{IEEEtran}
\usepackage{amsmath,amsfonts}
\usepackage{algorithmic}
\usepackage{array}
\usepackage{textcomp}
\usepackage{stfloats}
\usepackage{url}
\usepackage{verbatim}
\usepackage{graphicx}
\usepackage{algorithm,algorithmic,amsmath,amssymb,amsthm,balance,booktabs}
\usepackage{breakurl,color,colortbl,cuted,diagbox,enumitem}
\usepackage{float,graphicx,lipsum,lscape,makecell,mathtools,microtype,mparhack}
\usepackage{multirow,outlines,pifont,rotating,tabularx,tikz,times,url,xcolor}
\usepackage{subcaption}
\usepackage[colorlinks,urlcolor=black,linkcolor=black,citecolor=black]{hyperref}

\newcommand{\eg}{e.\,g.\,, }

\newenvironment{myquote}%
  {\it \list{}{\leftmargin=0.05in\rightmargin=0.05in}\item[]}%
  {\endlist}

\def\BibTeX{{\rm B\kern-.05em{\sc i\kern-.025em b}\kern-.08em
    T\kern-.1667em\lower.7ex\hbox{E}\kern-.125emX}}
\usepackage{balance}

\begin{document}

\title{A Wide Evaluation of ChatGPT \\ on Affective Computing Tasks}

\author{Mostafa M.\ Amin, Rui Mao, Erik Cambria, Bj\"orn W.\ Schuller

\IEEEcompsocitemizethanks{
\IEEEcompsocthanksitem  Mostafa M. Amin is with the Chair of Embedded Intelligence for Health Care and Wellbeing, University of Augsburg, 86159 Augsburg, Germany, and also with AI R\&D Team,  SyncPilot GmbH, 86156 Augsburg, Germany. E-mail: \href{mailto:mostafa.mohamed@uni-a.de}{mostafa.mohamed@uni-a.de}.
\IEEEcompsocthanksitem Rui Mao and Erik Cambria are with the School of Computer Science and Engineering, Nanyang Technological University, 639798 Singapore, Singapore.
\IEEEcompsocthanksitem Bj\"orn W. Schuller is with GLAM -- The group on Language, Audio, and Music, Imperial College London, SW7 2BX London, U.K., and also with the Chair of Embedded Intelligence for Health Care and Wellbeing, University of Augsburg, 86159 Augsburg, Germany.%
}
}


\markboth{Preprint}{How to Use the IEEEtran \LaTeX \ Templates}

\IEEEtitleabstractindextext{%
\begin{abstract}
With the rise of foundation models, a new artificial intelligence paradigm has emerged, by simply using general purpose foundation models with prompting to solve problems instead of training a separate machine learning model for each problem.
Such models have been shown to have emergent properties of solving problems that they were not initially trained on.
The studies for the effectiveness of such models are still quite limited.
In this work, we widely study the capabilities of the ChatGPT models, namely GPT-4 and GPT-3.5, on 13 affective computing problems, namely
aspect extraction, aspect polarity classification, opinion extraction, sentiment analysis, sentiment intensity ranking, emotions intensity ranking, suicide tendency detection, toxicity detection, well-being assessment, engagement measurement, personality assessment, sarcasm detection, and subjectivity detection.
We introduce a framework to evaluate the ChatGPT models on regression-based problems, such as intensity ranking problems, by modelling them as pairwise ranking classification.
We compare ChatGPT against more traditional NLP methods, such as end-to-end recurrent neural networks and transformers.
The results demonstrate the emergent abilities of the ChatGPT models on a wide range of affective computing problems, where GPT-3.5 and especially GPT-4 have shown strong performance on many problems,
particularly the ones related to sentiment, emotions, or toxicity.
The ChatGPT models fell short for problems with implicit signals, such as engagement measurement and subjectivity detection.
\end{abstract}

}

\maketitle

\section{Introduction}

\label{sec:introduction}

With the introduction
of foundation models~\cite{Bommasani21-Foundation,Zhau23-FoundationHistory},
a new paradigm to utilise machine learning models was introduced 
in Natural Language Processing (NLP).
The paradigm relies on the emerging capabilities of Large Language Models (LLMs)~\cite{Wei22-EmergLLMs} to perform more complex tasks with scaling.
Instead of training specialised models for specific problems, a large general foundation model would be trained once with general knowledge and this would be utilised (via prompting) in many other problems later on.
This paradigm has been introduced by language models as few-shot learners~\cite{Brown2020-LMFS}, and popularised with the launch of the groundbreaking ChatGPT foundation models, namely GPT-3.5~\cite{Ouyang22-InstructGPT} and its superior GPT-4~\cite{OpenAI23-GPT4}, which utilised techniques like Reinforcement Learning with Human Feedback (RLHF)~\cite{Ouyang22-InstructGPT}.
The emerging capabilities of ChatGPT are being studied in several domains, mostly either related to general artificial intelligence~\cite{Bubeck23-SparksAGI},
or traditional NLP problems, namely neural machine translation~\cite{Hendy23-ChatGPT-NMT}, and named entity recognition (NER)~\cite{Li23-Prompt-MNER}.

We examined the emerging capabilities of ChatGPT in affective computing in an early evaluation in a previous work~\cite{Amin23-WAC}, where we evaluated the performance of ChatGPT on three affective computing problems, namely suicide tendency detection, big-five personality assessment, and sentiment analysis. The study has shown interesting initial results confirming the emerging capabilities of ChatGPT, with results comparable to classical NLP methods such as Bag-of-Words (BoW)~\cite{Bishop2006} and Word2Vec~\cite{Mikolov13-Word2Vec}, whereas worse than fine-tuned language models like RoBERTa~\cite{Liu19-RoBERTa}. This study was performed on an early stage of ChatGPT, when the API for it was not yet released. Consequently, it was done manually on a small evaluation set. Additionally, given that prompt engineering is an emerging field, the prompting was less developed. Last but not least, the former study of ChatGPT evaluations focused on classification-based affective computing tasks, where regression-based tasks were not evaluated.

In this work, we aim to extend these evaluations beyond the limitations of the previous work~\cite{Amin23-WAC}, 
namely by exploring better prompting, bigger evaluation, handling of different learning tasks, and the wider scope of affective computing problems~\cite{Cambria2017-ACSA, Schuller13-CPE}. We examine a total of 34 setups for 13 affective computing problems. The problems are aspect extraction, aspect polarity classification, opinion extraction, sentiment analysis, sentiment intensity ranking, emotions intensity ranking, suicide tendency detection, toxicity detection, well-being assessment, engagement measurement, personality assessment, sarcasm detection, and subjectivity detection. The evaluation of these tasks is conducted due to their recognised prominence in the field of affective computing. These tasks collectively encompass various dimensions of understanding affective language and are widely acknowledged as pivotal in this domain.

Our evaluation yields the following findings: ChatGPT models excel in sentiment-related tasks, outperforming supervised baselines in opinion extraction, emotions, and sentiment intensity ranking. They particularly excel in identifying extremely negative emotions, notably in well-being assessment and toxicity detection, likely due to safety-focused training. However, they struggle with implicit signal tasks such as engagement measurement, personality assessment, sarcasm detection, and subjectivity detection. 

The contributions of this paper are as follows:
\begin{enumerate}
\item Executing a wide evaluation of the performance of the ChatGPT models, namely GPT-3.5 and GPT-4, on a wide range of affective computing problems.
\item Introducing a prompting framework that can be applied to LLMs for a wide range of affective computing problems.
\item Introducing a framework for the conversion of a regression task into a pairwise ranking classification task for the purpose of evaluating ChatGPT models.
\end{enumerate}

The paper is organised as follows: we discuss related work in the next section; afterwards, we introduce our method; then, we present and discuss the experimental results; finally, we provide concluding remarks.

\section{Related Work}
We introduce related works with a primary focus on the evaluation of foundation models in affective computing or NLP tasks.
Affective computing is an important research domain in NLP, including diverse tasks, such as sentiment analysis~\cite{mao2021bridging,cambria2022senticnet}, emotion detection~\cite{li2023skier,mao2023biases}, sarcasm detection~\cite{jayaraman2022sarcasm,yue2023knowlenet}, personality analysis~\cite{zhu2023pead}, mental health analysis~\cite{rastogi2022stress,han2022hierarchical}, figurative language processing~\cite{mao2018word,mao2023metaproonline}, and more. 
\cite{Zhang23-SentAI} evaluates ChatGPT on various sentiment analysis-related tasks, including aspect extraction.
\cite{Wang23-ICG-sentiment} evaluates ChatGPT on sentiment analysis, while~\cite{Ji23-ICG-personality} evaluates ChatGPT's ability to predict personality.
%
The works of~\cite{Kocon23-ChatGPT-Jack,Zhong23-ChatGPT-understanding,Qin23-chatGPT-general-solver} explore the capabilities of ChatGPT on a wide range of NLP problems including sentiment analysis and emotion recognition, and other NLP tasks like NER and text summarising.
Furthermore, we evaluated the fusion capabilities of ChatGPT with traditional NLP methods in \cite{Amin23-CCR-Fusion}.

The aforementioned ChatGPT models assessments concentrated on a confined subset of tasks within the domain of affective computing, notably sentiment analysis and emotion detection.
Nonetheless, considering the expansive scope of the affective computing field, there exists merit in conducting a comprehensive evaluation of the proficiency exhibited by the ChatGPT models within this domain.
This is motivated by the recognition that distinct affective computing tasks encapsulate intricate emotional nuances, demanding ChatGPT models to adeptly discern them across diverse application contexts.
A comprehensive evaluation encompassing a wider array of tasks also serves as a means to uncover any inherent biases that might manifest in the performance of the ChatGPT models.

On the other hand, the latest survey on the assessments of the ChatGPT models has unveiled a notable discrepancy in findings, stemming from the employment of diverse prompting strategies~\cite{mao2023gpteval}. Moreover, the evaluation process has scarcely encompassed regression tasks, due to the difficulty of prompting objective scores from LLMs in different task setups. This highlights the imperative of adopting a unified prompting framework to counteract the biases introduced by distinct affective computing evaluation tasks. Furthermore, the establishment of an efficacious evaluation methodology for assessing LLMs on regression tasks holds informative value for forthcoming research endeavours.

\section{Methodology}
\label{sec:method}




\subsection{Utilising ChatGPT}
\label{subsec:prompts}

We utilise the API provided by OpenAI to query ChatGPT\footnote{\url{https://platform.openai.com/docs/guides/gpt/chat-completions-api}}, using `gpt-3.5-turbo-0301' for GPT-3.5 and `gpt-4-0314' for GPT-4.
For each problem and target label combination, we construct a prompt for it, which is used as the system prompt.
For each given example in the evaluation data, we send two messages to ChatGPT, namely, the system prompt of the problem and a user message that includes only the input text of the example.
The assistant response acquired by ChatGPT is then the answer for the example. We exclude the examples that did not have simple parsing criteria.
Unlike the previous work~\cite{Amin23-WAC}, the extra specificity in the system prompt made the vast majority of examples easy to parse.
%

\subsection{Prompting}

\noindent
The prompt design follows a general pattern:
\begin{enumerate}
    \item Define the role of the assistant as an expert in the given problem, and add a problem description.
    \item Define the prediction task, namely, by specifying the type of labels and stating that the user will give an input and the assistant should reply to that by predicting the label for that example.
    \item Emphasise the answer format in bullet points of the exact requested format, and also the \emph{not} allowed formats. We do not allow it to explain the reasoning behind the decision, or mentioning statements like \emph{My guess is $\cdots$}.
    
\end{enumerate}


\subsubsection{\textbf{Word-level labels prompts}}

\begin{enumerate}
\item \textbf{Aspect extraction with polarity classification prompt}
\begin{myquote}
You are an aspect-based sentiment analysis expert,
you will be given a sentence by the user and you will list all the aspect words target objects.
List the words in bullet points.
The aspect targets are objects that are classified by a corresponding one of four sentiment targets: positive, negative, neutral, and conflict.
It is possible that a word has no target, which is defined as a background target. 
Use the following format:\\
* You will output a list of words in bullet points.\\
* Each bullet point will be on the form: ``word'' is target.\\
* The target is one of the four targets, do not report background targets.\\
* You will not mention any other text like ``My guess is ...'' or ``I think ...''.\\
* If all words have background target, then you return the word ``BACKGROUND'' without any bullet points.
\end{myquote}

\item \textbf{Opinion extraction prompt}
\begin{myquote}
You are an aspect-based sentiment analysis expert,
you will be given a sentence by the user that contains aspect words objects.
Your task is to list all the sentiment opinionated words / expressions, that are corresponding to the aspect in the text (if any).
You just need to list the words/expression in bullet points without classifying them.
There will be many words without sentiment, these should not be listed.
Use the following format:\\
* You will output a list of words in bullet points.\\
* Each bullet point will be on the form (without quotations): ``* expression''\\
* You should mention words that are explicitly in the text.\\
* You will not mention implied sentiment.\\
* You should mention the words exactly how they are written in the input, even if they have typos.\\
* You will not mention any other text like ``My guess is ...'' or ``I think ...''.\\
* If all words have no sentiment, then you respond with the word ``BACKGROUND'' without any bullet points.
\end{myquote}
\end{enumerate}

\subsubsection{\textbf{Sentence-level labels prompts}}

The following text is appended at the end of each of the following prompts.

\begin{myquote}
Use the following format:\\
* You are only allowed to answer ``\{label$_A$\}'' or ``\{label$_B$\}''.\\
* Don't write an explanation of the answer.\\
* Don't write things like ``My guess is...'', or ``I think ...''. Just write \{label$_A$\} or \{label$_B$\}, but nothing else.
\end{myquote}

\begin{enumerate}
\item \textbf{Sentiment analysis prompt}
\begin{myquote}
You are an expert at sentiment analysis.
Given a text by the user, analyze the sentiment of the text if it is `positive' or `negative'.
You are not allowed to answer `neutral', try to narrow it down to `positive' and `negative'.
\end{myquote}

\item \textbf{Sentiment intensity ranking prompt}
\begin{myquote}
You are an expert at sentiment analysis.
Given a pair of texts A and B from the user,
you will output which text expresses more positive sentiment.
\end{myquote}

\item \textbf{Emotion intensity ranking prompt}
\begin{myquote}
You are an expert at emotion analysis.
Given a pair of texts A and B from the user,
you will output which text expresses higher intensity of the \{emotion\} emotion.
\end{myquote}

\item \textbf{Suicide detection prompt}
\begin{myquote}
You are an expert at psyche analysis.
Given a text by the user, solve the binary classification of analysing if the text expresses a tendency for suicide.
\end{myquote}

\item \textbf{Toxicity detection prompt}
\begin{myquote}
You are an expert at toxicity analysis.
Assume that we have the capability of analysing 6 toxicity traits.
``toxic'', ``severe toxic'', ``obscene'', ``threat'', ``insult'', ``identity hate''.
Your task is to make binary classification for the trait \{trait\}, and not the remaining traits.
Given a text by the user, estimate if the given text displays the trait \{trait\} or not.
\end{myquote}

\item \textbf{Well-being assessment prompt}
\begin{myquote}
You are an expert at psyche analysis.
Given a text by the user, estimate if the given text talks about a stress-related topic, or expresses emotional stress be it implicit or explicit.
\end{myquote}

\item \textbf{Engagement measurement prompt}
\begin{myquote}
You are an expert at social media analysis.
Given a pair of texts A and B representing tweets, estimate which text is more engaging.
You will achieve this by estimating which text is more viral,
by estimating which one has a higher number of retweets.
\end{myquote}

\item \textbf{Personality assessment prompt}
\begin{myquote}
You are an expert at the big-five personality traits assessment.
Given a pair of texts A and B from the user,
you will output which text expresses higher intensity of the \{trait\} trait, from the big-five OCEAN personality traits.
\end{myquote}

\item \textbf{Sarcasm detection prompt}
\begin{myquote}
You are an expert at sarcasm analysis.
Given a text by the user, estimate if the given text is sarcastic or not.
\end{myquote}

\item \textbf{Subjectivity detection prompt}
\begin{myquote}
You are an expert at language and sentiment analysis.
The user will give you a text, your task is to make a binary classification on the text,
if the given text is opinionated / subjective / biased, or if it is non-opinionated / objective / descriptive / factual.
Please note that this is about ``how'' the text is described and not ``what'' it describes,
so the text can still ``objectively'' describe a fictional story with some emotional terms.
\end{myquote}

\end{enumerate}

\subsection{Pairwise comparison to solve regression tasks}
\label{subsec:regression}

Querying and prompting ChatGPT models for classification tasks can be readily accomplished by instructing them to select a suitable label from the predefined label set of classification tasks. Nonetheless, the challenge arises when transitioning to regression tasks, as querying an objective score from the ChatGPT models becomes intricate due to the variations in scales and the inherent subjectivity associated with dataset annotation criteria. Thus, we evaluate ChatGPT on regression problems by modelling regression problems as a pairwise ranking classification problem.
In other words, given $N$ regression labels in the evaluation data, we can remodel the problem into a classification problem by sampling $M$ pairs $(a, b)$, and solve the binary classification problem `\emph{is $y_a > y_b$?}'. A crucial aspect of this framework is how to sample the $M$ pairs to be as few as possible whilst being representative. We employ the small-world graph generation algorithm~\cite{Chen16-OCP}, which is a densely connected graph without having a very high number of connections. In our setup we sample $M=4N$ pairs.

This modelling can be further applied for predictions not just evaluations. A prediction procedure would utilise successive halving of a range of answers by comparisons with the median item within the remaining range, until the range has a small width, similar to the binary search algorithm. However, we do not use this prediction mechanism in this work.
\subsection{Compared Model}
\label{subsec:models}

To evaluate the performance of the ChatGPT models, we compare them with a Recurrent Neural Network (RNN)-based framework that consists of a single bidirectional LSTM layer (with $L$ units in each direction) followed by $N$ fully-connected layers with $U$ units (with ReLU activations), then a final prediction layer.
The final layer uses sigmoid activation in most setups, except for aspect-based problems (softmax for multi-class classification), sentiment ranking (tanh), and engagement measurement (ReLU). This framework leverages different features that will be introduced in the following section.

Adam~\cite{Diederik15-Adam} is employed as an optimisation algorithm, with a learning rate $\alpha$.
The loss function is crossentropy for classification tasks~\cite{Bishop2006}, and Mean Absolute Error (MAE) for regression tasks~\cite{Bishop2006}.
For problems with imbalanced datasets, namely toxicity detection and aspect extraction,
we make use of a weighting parameter $\lambda$ that discounts the weight of the `0/negative' class (typically the over-represented class).
%
In order to tune the hyperparameters $\lambda, \alpha, N, U, L$, we opt to utilise the hyperparameter optimisation toolkit SMAC~\cite{Lindauer22-SMAC} to select the best hyperparameters for each problem.
The hyperparameter space is $N \in [0,1]$, $U \in [64, 512]$ (log-sampled), $L \in [16, 64]$ (log-sampled), $\alpha \in [10^{-6}, 10]$ (log-sampled),
and $\lambda \in [10^{-3}, 1]$ (log-sampled).
We sample 25 combinations for each problem.
In each training run, we train for a total of 300 epochs with early stopping (30 epochs).
The model with the best validation score is the one used for testing.

\subsection{Text Features}
We utilise two different textual representations of the input texts and training an RNN on top of that.
The representations are the raw text as sequence of word IDs (we call that the end-to-end (E2E) approach), and RoBERTa features~\cite{Liu19-RoBERTa}
In the E2E approach, the IDs are limited to the most common 2,000 words in the training set of a given problem, and we train embeddings of dimension 128 for that, jointly with the rest of the model.

\textbf{RoBERTa Language Model}
We employ the RoBERTa language model to extract features for a given sentence,
by basically running the RoBERTa-base model\footnote{Acquired on 28.07.2023 from \url{https://huggingface.co/docs/transformers/model_doc/roberta}} on a given sentence to acquire a sequence of features, corresponding to features of the subwords.
Each subword is represented by a feature vector of size 768.
The RoBERTa-base model~\cite{Liu19-RoBERTa} is the smaller variant of the RoBERTa architecture, which is based on the BERT transformer architecture~\cite{Devlin2019-BERT}.
For word-based labelling in the aspect-related problems, we give all subwords the same label as the corresponding word at training, while using the prediction of the first subword of a word as the corresponding word prediction at evaluation similar to~\cite{Devlin2019-BERT}.

\subsection{Datasets}
\label{subsec:datasets}

We employ 13 datasets for the 13 evaluation tasks in the affective computing domain. The summary of the statistics of our employed datasets is given in Table~\ref{tab:data-stats}. For training and validation, we make use of the original splits provided with the datasets, or we split them otherwise\footnote{Splits procedures will be provided on: \url{https://github.com/mostafa-mahmoud/chatgpt-wide-evaluation} upon acceptance.}. For benchmarking, we always downsample the testing set if the original testing set is not small enough, due to the very tight restrictions on scaled usage of GPT-4.

\begin{table}[!t]
    \centering
    \begin{tabular}{c|c||c|c|c}

\multicolumn{2}{c||}{Problem}  &  Train & Dev & Test \\
\hline
\multirow{3}{*}{\rotatebox{90}{ABSA}} 
& res14 &  2,436  &  608  &  800\\ \cline{2-5}
& lap14 &  2,439  &  609  &  800\\ \cline{2-5}
& res15 &  1,052  &  263  &  685\\ 
\hline
\multicolumn{2}{c||}{Sentiment Analysis}  &  100,000  &  10,000  &  2,500\\ 
\hline
\multicolumn{2}{c||}{Sentiment Ranking} &  1,000  &  300  &  365\\ 
\hline
\multirow{4}{*}{\rotatebox{90}{Emotion}} 
& Sadness &  786  &  74  &  673\\ 
& Joy &  823  &  79  &  714\\ 
& fear &  1,147  &  110  &  995\\ 
& Anger &  857  &  84  &  760\\ 
\hline
\multicolumn{2}{c||}{Suicide}  &  23,398  &  5,611  &  2,345\\ 
\hline
\multicolumn{2}{c||}{Toxicity}  &  30,000  &  6,864  &  959\\ 
\hline
\multirow{4}{*}{\rotatebox{90}{Well-be.}} 
& Reddit bodies  &  1,511  &  458  &  935\\ 
& Reddit titles  &  3,538  &  996  &  998\\ 
& Twitter denoised &  851  &  400  &  800\\ 
& Twitter full  &  5,900  &  1,500  &  1,500\\ 
\hline
\multicolumn{2}{c||}{Engagement}  &  30,037  &  5,000  &  4,000\\ 
\hline
\multicolumn{2}{c||}{Personality}  &  5,992  &  2,000  &  1,996\\ 
\hline
\multicolumn{2}{c||}{Sarcasm}  &  18,709  &  4,000  &  4,000\\ 
\hline
\multicolumn{2}{c||}{Subjectivity}  &  6,000  &  2,000  &  2,000\\ 
    \end{tabular}
    \caption{Datasets sizes statistics; these are shown on our selection of the data, after we downsample (and possibly split) some of the datasets. ABSA denotes aspect-based sentiment analysis that includes aspect extraction, opinion extraction and aspect polarity classification subtasks.}
    \label{tab:data-stats}
    \vspace{-0.3cm}
\end{table}

\noindent \textbf{The Aspect-based Sentiment Analysis Datasets} are from the SemEval 2014~\cite{pontiki2014semeval} and SemEval 2015~\cite{pontiki2015semeval} shared tasks. The shared tasks sourced data from laptop and restaurant reviews, termed lap14, res14, and res15 in the later result table. We employ the split of training, validation, and testing sets from the work of~\cite{chen2020racl}. The dataset contains three sets of word annotations labels for aspect extraction, aspect polarity prediction, and opinion extraction tasks.

\noindent \textbf{The Sentiment Analysis Dataset} is the Twitter140 dataset~\cite{Go2009-Twitter} which consists of tweets and their corresponding sentiment negative or positive labels.
Given the extremely small size of the provided testing set, we downsample and split the large training set into the Train/Dev/Test sets we use.

\noindent \textbf{The Sentiment Intensity Ranking Dataset} is from the SemEval-2017 Shared Task 5~\cite{cortis2017semeval}. The dataset sourced data from two domains, namely, microblog messages and news headlines. The sentiment intensity scores are within $[-1,1]$. Our experiments are conducted on the microblog data.

\noindent \textbf{The Emotion Intensity Dataset} is from the WASSA-2017 Shared Task on Emotion Intensity (EmoInt)~\cite{mohammad2017wassa}. The dataset includes four emotions, namely joy, sadness, fear, and anger and emotion intensity scores, ranging within $[0, 1]$.

\noindent \textbf{The Suicide Tendency Detection Dataset} is from the work of~\cite{Desu22-Suicide}, which is a dataset collected from Reddit under depression and teenagers Subreddits, for positive and negative labels, respectively.

\noindent \textbf{The Toxicity Detection Dataset} is from the Toxic Comment Classification Challenge\footnote{\url{https://kaggle.com/competitions/jigsaw-toxic-comment-classification-challenge}}. The dataset encompasses an extensive collection of Wikipedia comments that have been meticulously annotated by humans. These instances of toxicity encompass various categories, namely: toxicity, severe toxicity, obscenity, threats, insults, and identity-based hate.
We split the training part into Train/Dev sets, and downsample
the negative class because it is extremely over-represented in comparison.

\noindent \textbf{The Well-being Assessment Dataset} is from the work of~\cite{rastogi2022stress}, where the data were from Reddit and Twitter, with two benchmarks on each.
For the (Combi) Reddit benchmark, we make use of the bodies of the posts, instead of the titles.
Binary labels are used in this dataset, indicating stress-negative and stress-positive text.

\noindent \textbf{The Engagement Measurement Dataset} is from Kaggle TEDTalks Tweets\footnote{\url{https://www.kaggle.com/datasets/thedevastator/social-media-interactions-on-tedtalks-dataset}}. The dataset was obtained from Twitter and pertains to discussions associated with TEDTalks. The dataset contains tweet content, and the number of likes. This allows us to conduct a comparative analysis, discerning the relative favourability of different content.
We use the $\log_{10}(\text{number of retweets} + 1)$ as the target label, since it represents the labels distribution effectively.

\noindent \textbf{The Personality Assessment Dataset} is the First Impressions dataset~\cite{Ponce16-ChaLearn}, from which we use the transcripts of personality annotated videos.
The labels are expressed by the big-five personality model (Openness, Conscientiousness, extraversion, agreeableness, and neuroticism), with labels within $[0,1]$.

\noindent \textbf{The Sarcasm Detection Dataset} is from the work of~\cite{misra2023Sarcasm} (Version 1). The data were sourced from \textit{TheOnion} and \textit{HuffPost} news headlines, associated with binary labels, indicating if a news headline is sarcastic or not.

\noindent \textbf{The Subjectivity Detection Dataset} is from the work of~\cite{pang2004sentimental}. The data were sourced from movie reviews from \textit{Rotten Tomatoes}, associated with binary labels, indicating if a sentence or a piece of text is subjective or objective.

\section{Results}
\label{sec:experiments}

\begin{table*}[!t]
    \centering
    \begin{tabular}{c|c||l|l|l|l||l|l|l|l}
    \hline
    \multicolumn{2}{c||}{\multirow{2}{*}{Dataset}} & \multicolumn{4}{c||}{Accuracy [\%]} & \multicolumn{4}{c}{UAR [\%]} \\ \cline{3-10}
    \multicolumn{2}{c||}{} & GPT-3.5 & E2E & RoBERTa & GPT-4 & GPT-3.5 & E2E & RoBERTa & GPT-4 \\
    \hline

\hline 
\multirow{3}{*}{\parbox[c]{1cm}{\centering Aspect \\ Extraction}} 
&   res14  &  $        86.95   $  &  $        81.73^{**} $  &  $\textbf{92.00}^{**} $  &  $        71.50^{**} $  &  $        76.18   $  &  $        81.18^{**} $  &  $\textbf{91.54}^{**} $  &  $        73.23^{**} $  \\
&   lap14  &  $        84.60   $  &  $        78.22^{**} $  &  $\textbf{87.19}^{**} $  &  $        70.32^{**} $  &  $        77.62   $  &  $        82.21^{**} $  &  $\textbf{90.64}^{**} $  &  $        75.94      $  \\
&   res15  &  $\textbf{84.57}   $  &  $        81.28^{**} $  &  $        73.02^{**} $  &  $        70.05^{**} $  &  $        78.56   $  &  $        72.28^{**} $  &  $\textbf{85.13}^{**} $  &  $        73.81^{**} $  \\

\hline 
\multirow{3}{*}{\parbox[c]{1cm}{\centering Aspect \\ Polarity}} 
&   res14  &  $        85.13   $  &  $\textbf{86.10}^*    $  &  $        71.85^{**} $  &  $        69.30^{**} $  &  $        49.96   $  &  $        49.72      $  &  $\textbf{58.09}^*    $  &  $        48.26      $  \\
&   lap14  &  $        82.23   $  &  $        72.57^{**} $  &  $\textbf{90.22}^{**} $  &  $        67.63^{**} $  &  $        47.37   $  &  $        45.10      $  &  $\textbf{58.23}^{**} $  &  $        44.54^{**} $  \\
&   res15  &  $        82.38   $  &  $        79.08^{**} $  &  $\textbf{84.31}^{**} $  &  $        67.51^{**} $  &  $        47.79   $  &  $        36.84^{**} $  &  $\textbf{48.96}      $  &  $        44.27      $  \\

\hline 
\multirow{3}{*}{\parbox[c]{1cm}{\centering Opinion \\ Extraction}} 
&   res14  &  $        91.04   $  &  $        81.61^{**} $  &  $\textbf{93.26}^{**} $  &  $        80.93^{**} $  &  $        77.28   $  &  $        80.87      $  &  $        80.06      $  &  $\textbf{83.02}^*    $  \\
&   lap14  &  $\textbf{89.43}   $  &  $        74.33^{**} $  &  $        73.81^{**} $  &  $        76.90^{**} $  &  $        73.51   $  &  $        66.43^*    $  &  $        76.73      $  &  $\textbf{77.42}      $  \\
&   res15  &  $\textbf{89.32}   $  &  $        79.42^{**} $  &  $        89.16      $  &  $        78.10^{**} $  &  $        76.90   $  &  $        68.59      $  &  $\textbf{77.31}      $  &  $        77.19      $  \\

\hline
\multicolumn{2}{c||}{Sentiment Analysis}  &  $        80.54   $  &  $        78.87      $  &  $\textbf{88.74}^{**} $  &  $        84.09^{**} $  &  $        79.93   $  &  $        78.84      $  &  $\textbf{88.75}^{**} $  &  $        83.69^{**} $  \\

\hline
\multicolumn{2}{c||}{Sentiment Ranking}  &  $        69.30   $  &  $        70.88      $  &  $        72.37      $  &  $\textbf{73.21}^{**} $  &  $        68.69   $  &  $        70.83      $  &  $        72.41^*    $  &  $\textbf{73.08}^{**} $  \\

\hline
\multirow{4}{*}{\parbox[c]{1cm}{\centering Emotion \\ Ranking}} 
&   Joy  &  $        74.07   $  &  $        66.49^{**} $  &  $        75.41      $  &  $\textbf{78.46}^{**} $  &  $        74.29   $  &  $        66.51^{**} $  &  $        75.40      $  &  $\textbf{78.63}^{**} $  \\
&   Fear  &  $        72.76   $  &  $        68.65^{**} $  &  $\textbf{76.83}^{**} $  &  $        73.96      $  &  $        72.86   $  &  $        68.65^{**} $  &  $\textbf{76.83}^{**} $  &  $        74.09      $  \\
&   Anger  &  $        72.12   $  &  $        67.63^{**} $  &  $        73.47      $  &  $\textbf{75.58}^{**} $  &  $        72.09   $  &  $        67.60^{**} $  &  $        73.46      $  &  $\textbf{75.49}^{**} $  \\
&   Sadness  &  $        78.19   $  &  $        72.41^{**} $  &  $        76.06      $  &  $\textbf{78.55}      $  &  $        78.22   $  &  $        72.40^{**} $  &  $        76.05      $  &  $\textbf{78.58}      $  \\

\hline
\multicolumn{2}{c||}{Suicide Detection}  &  $        89.46   $  &  $        84.75^{**} $  &  $\textbf{98.43}^{**} $  &  $        93.46^{**} $  &  $        89.44   $  &  $        85.32^{**} $  &  $\textbf{98.46}^{**} $  &  $        93.32^{**} $  \\

\hline
\multirow{6}{*}{Toxicity}
&   Toxic  &  $        87.37   $  &  $        81.85^{**} $  &  $        85.23      $  &  $\textbf{89.29}      $  &  $        87.19   $  &  $        82.75^{**} $  &  $        86.01      $  &  $\textbf{89.65}^*    $  \\
&   Severe toxic  &  $        66.55   $  &  $\textbf{87.65}^{**} $  &  $        80.07^{**} $  &  $        75.52^{**} $  &  $        80.65   $  &  $        82.48      $  &  $        84.78^*    $  &  $\textbf{85.29}^{**} $  \\
&   Obscene  &  $        83.45   $  &  $        85.40      $  &  $        84.83      $  &  $\textbf{88.16}^{**} $  &  $        83.48   $  &  $        86.62^*    $  &  $        86.60^*    $  &  $\textbf{86.78}^{**} $  \\
&   Threat  &  $        70.59   $  &  $        94.05^{**} $  &  $\textbf{95.54}^{**} $  &  $        91.99^{**} $  &  $        80.12   $  &  $        73.99      $  &  $        87.43^*    $  &  $\textbf{91.51}^{**} $  \\
&   Insult  &  $        80.14   $  &  $        84.65^{**} $  &  $\textbf{87.25}^{**} $  &  $        80.70      $  &  $        83.21   $  &  $        84.64      $  &  $\textbf{89.28}^{**} $  &  $        84.89      $  \\
&   Identity hate  &  $        66.82   $  &  $        90.52^{**} $  &  $\textbf{90.98}^{**} $  &  $        82.66^{**} $  &  $        78.61   $  &  $        81.61      $  &  $        86.16^{**} $  &  $\textbf{87.88}^{**} $  \\

\hline
\multirow{4}{*}{Well-being}
&    Reddit bodies  &  $        91.93   $  &  $        84.50^{**} $  &  $        89.88      $  &  $\textbf{93.33}      $  &  $        84.41   $  &  $        68.82^{**} $  &  $\textbf{86.16}      $  &  $        78.63^{**} $  \\
&    Reddit titles  &  $        80.61   $  &  $        86.60^{**} $  &  $\textbf{96.75}^{**} $  &  $        89.54^{**} $  &  $        80.05   $  &  $        85.62^{**} $  &  $\textbf{96.65}^{**} $  &  $        89.77^{**} $  \\
&    Twitter denoised  &  $        60.53   $  &  $        43.36^{**} $  &  $\textbf{93.23}^{**} $  &  $        72.31^{**} $  &  $        65.05   $  &  $        45.45^{**} $  &  $\textbf{93.14}^{**} $  &  $        73.45^{**} $  \\
&    Twitter full  &  $        66.24   $  &  $        80.39^{**} $  &  $\textbf{84.39}^{**} $  &  $        75.25^{**} $  &  $        66.26   $  &  $        80.39^{**} $  &  $\textbf{84.39}^{**} $  &  $        75.25^{**} $  \\

\hline
\multicolumn{2}{c||}{Engagement} &  $        51.92   $  &  $        71.02^{**} $  &  $\textbf{79.18}^{**} $  &  $        54.15^{**} $  &  $        51.85   $  &  $        71.02^{**} $  &  $\textbf{79.19}^{**} $  &  $        53.80^{**} $  \\

\hline
\multirow{5}{*}{Personality}
&   Openness  &  $        50.11   $  &  $        58.36^{**} $  &  $\textbf{60.54}^{**} $  &  $        54.75^{**} $  &  $        50.60   $  &  $        58.36^{**} $  &  $\textbf{60.56}^{**} $  &  $        54.59^{**} $  \\
&   Conscient.    &  $        55.54   $  &  $        56.79      $  &  $\textbf{61.59}^{**} $  &  $        57.44^*    $  &  $        55.84   $  &  $        56.78      $  &  $\textbf{61.59}^{**} $  &  $        57.33      $  \\
&   Extraversion  &  $        53.55   $  &  $        56.51^{**} $  &  $\textbf{59.03}^{**} $  &  $        55.90^{**} $  &  $        53.38   $  &  $        56.51^{**} $  &  $\textbf{59.02}^{**} $  &  $        55.90^{**} $  \\
&   Agreeable.&  $        51.67   $  &  $        57.81^{**} $  &  $\textbf{58.12}^{**} $  &  $        54.04^{**} $  &  $        52.10   $  &  $        57.80^{**} $  &  $\textbf{58.14}^{**} $  &  $        54.05^*    $  \\
&   Neuroticism  &  $        48.94   $  &  $        58.60^{**} $  &  $\textbf{59.86}^{**} $  &  $        49.68      $  &  $        49.04   $  &  $        58.60^{**} $  &  $\textbf{59.86}^{**} $  &  $        49.73      $  \\
\hline
\multicolumn{2}{c||}{Sarcasm}  &  $        59.13   $  &  $        63.14^{**} $  &  $\textbf{90.66}^{**} $  &  $        66.66^{**} $  &  $        56.82   $  &  $        66.29^{**} $  &  $\textbf{90.70}^{**} $  &  $        69.45^{**} $  \\
\hline
\multicolumn{2}{c||}{Subjectivity}  &  $        59.56   $  &  $        87.28^{**} $  &  $\textbf{95.56}^{**} $  &  $        88.38^{**} $  &  $        58.59   $  &  $        87.19^{**} $  &  $\textbf{95.51}^{**} $  &  $        88.19^{**} $  \\

    \end{tabular}
    \caption{Classification accuracy and UAR scores for all the problems. The bold scores correspond to the best method on the given metric and problem. $^*, ^{**}$ correspond to scores with statistically significant results (with $p$-values $<5\,\%$ and $< 1\,\%$, respectively) as compared to GPT-3.5.}
    \vspace{0.1cm}
    \label{tab:performance}
\end{table*}

The main results of the experiments are shown in Table~\ref{tab:performance}, where we evaluate classification accuracy and Unweighted Average Recall (UAR), which is the unweighted average of the accuracy of classifying each class separately~\cite{Schuller13-TI2}.
For the time-series labels, we use micro-averaging for the measuring the performance.
For all results, we utilise a two-tailed randomised permutation test to check for the statistical significance of the differences in performance compared to GPT-3.5~\cite{Good94-PT}.


The results show that the RNN trained with RoBERTa features has the best performance in the majority of problems, for both metrics; in most of these, it is even significantly better than GPT-3.5.
Furthermore, GPT-4 comes at second best overall, with the best performance on some of the setups.
In the instances where GPT-3.5 is better than RoBERTa, it is with a difference that is not regarded as statistically significant, and it is often due to the fact that the RoBERTa-based approach prioritised improving UAR instead of accuracy. 
On the other hand, comparing GPT-3.5 to the simpler baseline E2E presents a different picture; E2E is sometimes significantly worse and sometimes significantly better than GPT-3.5.

For aspect extraction and aspect target predictions, RoBERTa has the best performance overall especially when considering UAR, followed by GPT-3.5.
However, E2E is better than GPT-3.5 (considering UAR only) to identify the aspect but not its polarity.
Surprisingly, GPT-4 has the worst performance only on this problem; upon inspection of few of the GPT-4 results, we found that despite its right identification of the aspect expressions, still it tends to include more surroundings words, which is probably a major reason behind this deterioration in performance.
For the opinion extraction problem, GPT-4 has the best performance, especially considering UAR, followed by RoBERTa, then GPT-3.5.

In sentiment related problems, RoBERTa and GPT-4 are far better than both GPT-3.5 and E2E.
GPT-3.5 and E2E have relatively close performance on sentiment-based problems.
The results of GPT-3.5 on sentiment analysis are similar to previous work~\cite{Amin23-WAC, Amin23-CCR-Fusion}. 

In emotions intensity ranking, GPT-4 is the best model (significantly better than GPT-3.5 and E2E), followed by RoBERTa, then GPT-3.5, then E2E (significantly worse than GPT-3.5).
An interesting observation is the strong performance of GPT-3.5 in identifying the emotion sadness; since ChatGPT models seem to be very competent in problems related to identifying negative emotions as we will elaborate. 

On psychology-related problems with extreme negative emotions, namely the detection of suicide tendency, well-being, and toxicity, the ChatGPT models have a strong performance in many cases that significantly outperforms E2E.
The results of the suicide-detection are consistent with previous work~\cite{Amin23-WAC,Amin23-CCR-Fusion}, where RoBERTa is significantly better than GPT-3.5.
Moreover, ChatGPT models seem to thrive with longer texts, as seen in the results of the well-being problem part Reddit bodies, which is the only part consisting of long texts of the full body of Reddit posts, where both GPT-4 and GPT-3.5 are achieving the best results;
on the shorter texts, RoBERTa is the best model by a wide margin.
Furthermore, the ChatGPT models are showing the best results for the toxicity problem.
This might come as non-surprising given the fact that the ChatGPT models from OpenAI generally are tuned to identify toxicity as part of applying safety policies\footnote{\url{https://openai.com/policies/usage-policies}}.
However, they are not as competent at specifying further the reason behind toxicity, given their inferiority compared to RoBERTa about the more specific toxicity labels.

For the tasks with more implicit or latent social signals, namely engagement measurement, personality assessment, sarcasm detection, and subjectivity detection, GPT-3.5 has very poor performance that is significantly worse than E2E in most cases and significantly worse than RoBERTa in all cases.
GPT-4 shows minor improvement over GPT-3.5 on these problems, where it slightly surpasses E2E only in the sarcasm and subjectivity detection problems.
Similar to previous work~\cite{Amin23-WAC,Amin23-CCR-Fusion}, the results of the personality from GPT-3.5 are the worst, even compared to the simple baseline BoW.
The results of the ChatGPT models on the engagement measurement problem are extremely poor, close to a random predictor (UAR 50\,\%).
In earlier experiments, we attempted to make GPT-3.5 solve different formulations of the engagement measurement problem; the first was binary classifying if the number of retweets is $< 10$, the second was classifying if the number of retweets is $< 100$, and the third is 3-class classifying the number of retweets $[0, 10), [10, 100), \text{ and } [100, \infty)$.
It still yielded the same poor results in all of them.


The effectiveness of the introduced regression evaluation technique in section~\ref{subsec:regression} is demonstrated by the results of the personality assessment problem.
In the previous work~\cite{Amin23-WAC,Amin23-CCR-Fusion}, personality assessment was evaluated as a binary classification problem, while the results of GPT-3.5 here are evaluated using the newly introduced pairwise ranking evaluation framework.
The consistency of the results and their relative order across the five independent labels indicate the effectiveness of the technique.
The same applies to the engagement measurement problem as mentioned earlier.

The issues of parsing results mentioned in the early evaluation~\cite{Amin23-WAC} are mostly resolved within this study.
This is due to the system prompt (which enforces instructions) introduced in the API version of ChatGPT models (which was not available at the early evaluation~\cite{Amin23-WAC}), and our more precise formulation of the prompts (see section~\ref{subsec:prompts}).
A crucial aspect is highlighting the possible answering formats while disallowing improper formats.
The redundancy is also helpful in this regard, 
since we specify the formatting in a general form once on the problem description (like~\cite{Amin23-WAC}), then, we specify it a second time in a precise manner in the notes bullet points (unlike~\cite{Amin23-WAC}).
This led to the behaviour that very few examples are not following the described format, unlike~\cite{Amin23-WAC}.
However, the issue still stands for the compound predictions, in particular, we faced the same parsing issues for the aspect-related problems because the predictions contained multiple labels, where ChatGPT models were improvising in the response format or modifying some of the mentioned input words.
As a result, the responses of ChatGPT models are not fully reliable for properly formatting complex predictions, only straight-forward predictions with a short answer.

\section{Conclusion}
\label{sec:conclusion}

In this paper, we evaluated ChatGPT models, namely GPT-3.5 and GPT-4, on 13 affective computing problems.
We prompted the ChatGPT models methodically for each problem, then, we used the API to retrieve predictions. We compared the ChatGPT models against two traditional natural language processing models, namely an end-to-end (E2E) Recurrent Neural Network (RNN), or extending the RNN by using the input features from the RoBERTa model. The results have shown that the general performance order of the models is championed by RoBERTa, followed by GPT-4, then, E2E and GPT-3.5 are the following rivals with no clear winner for all problems.
GPT-4 is mostly better than GPT-3.5,
except for the aspect extraction problem,
with a statistically significant difference in most cases. 

The ChatGPT models showed strong performance in sentiment-related problems, where GPT-3.5 is usually better than E2E, and GPT-4 tends to be often better than RoBERTa, \eg 
in the opinion extraction, and emotions and sentiment intensity ranking problems.
Both ChatGPT models have shown their strongest performance in problems identifying extremely negative emotions, especially for well-being assessment over long texts and toxicity detection; this is probably due to the extra training by OpenAI in an attempt to enforce safety policies. The problems that the ChatGPT models fell very short were problems with implicit signals, like engagement measurement, personality assessment, sarcasm detection, and subjectivity detection. GPT-3.5 has shown significantly worse performance than E2E for most of these, and GPT-4 was at best slightly better than E2E if not significantly worse for two of these problems. 
Future efforts can focus on reinforced prompt design, and a synergistic combination of the compared approaches.

\bibliographystyle{IEEEtran}
\bibliography{references}

\begin{IEEEbiography}[{\includegraphics[width=1in,height=1.25in,clip,keepaspectratio]{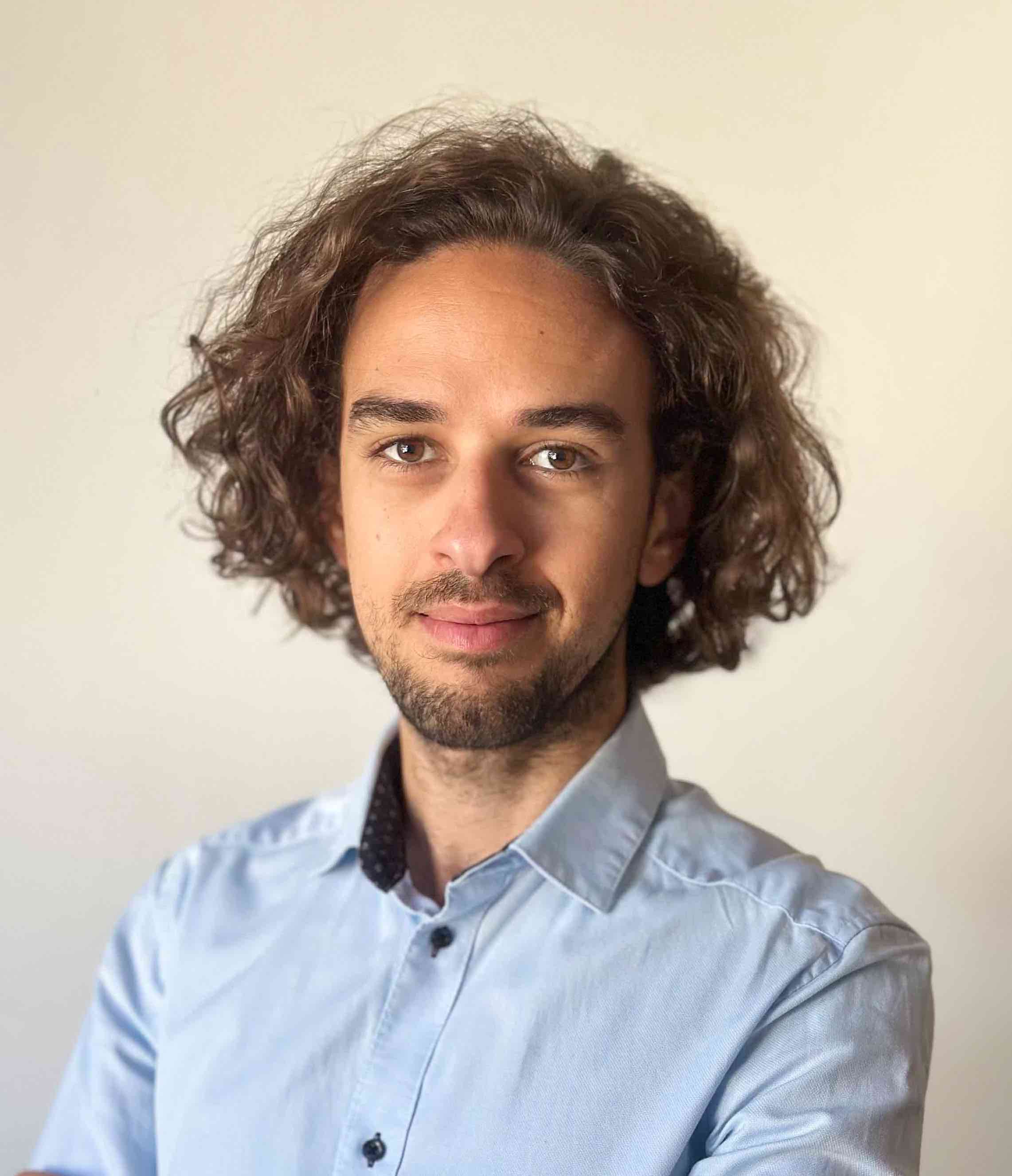}}]{Mostafa M. Amin}{\,}is currently working toward the Ph.D.~degree with the Chair of Embedded Intelligence for Health Care and Wellbeing with University of Augsburg, while working as Senior Research Data Scientist at SyncPilot GmbH in Augsburg, Germany. His research interests include Affective Computing, Audio and Text Analytics. He received a M.Sc.\ degree in Computer Science from the University of Freiburg, Germany. Contact him at \href{mailto:mostafa.mohamed@uni-a.de}{mostafa.mohamed@uni-a.de} 
\end{IEEEbiography}

\begin{IEEEbiography}[{\includegraphics[width=1in,height=1.25in,clip,keepaspectratio]{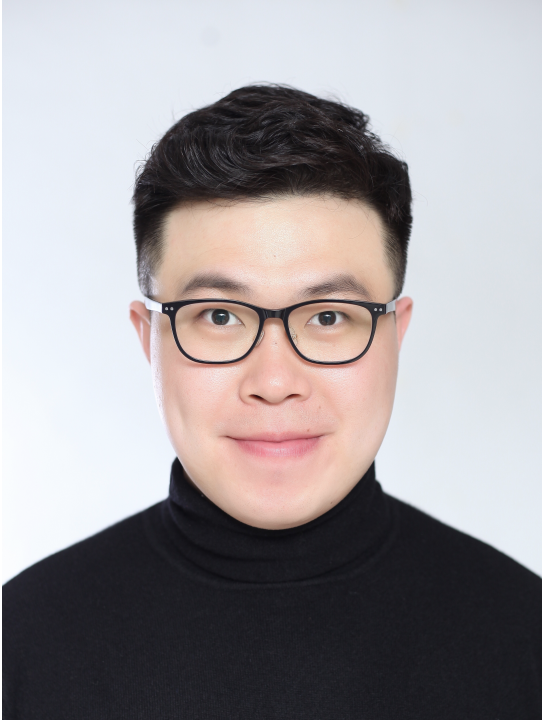}}]{Rui Mao}{\,}is a research fellow and lead investigator at Nanyang Technological University. He obtained his Ph.D. degree in Computing Science from the University of Aberdeen. His research interest lies at NLP, affective computing, and their applications in finance and cognitive science. He and his funded company (Ruimao Tech) have developed an end-to-end system (MetaPro) for computational metaphor processing and a neural search engine (wensousou.com) for searching Chinese ancient poems with modern language. Contact him at \href{mailto:rui.mao@ntu.edu.sg}{rui.mao@ntu.edu.sg}.
\end{IEEEbiography}

\begin{IEEEbiography}[{\includegraphics[width=1in,height=1.25in,clip,keepaspectratio]{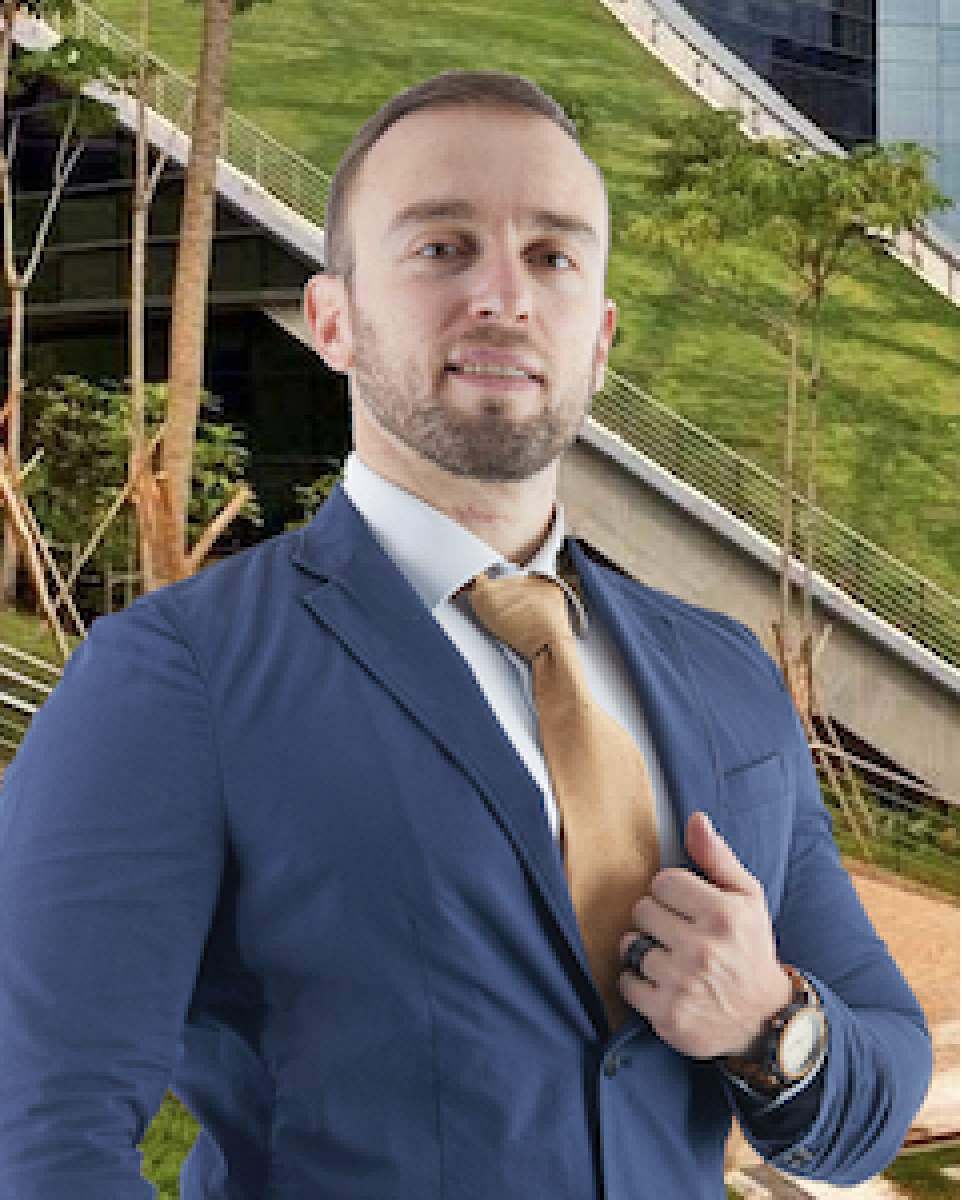}}]{Erik Cambria}{\,}is a professor of Computer Science and Engineering at Nanyang Technological University, Singapore. His research focuses on neurosymbolic AI for explainable natural language processing in domains like sentiment analysis, dialogue systems, and financial forecasting. He is an IEEE Fellow and a recipient of several awards, e.\,g., IEEE Outstanding Career Award, was listed among the AI's 10 to Watch, and was featured in Forbes as one of the 5 People Building Our AI Future. Contact him at \href{mailto:cambria@ntu.edu.sg}{cambria@ntu.edu.sg}.
\end{IEEEbiography}

\begin{IEEEbiography}[{\includegraphics[width=1in,height=1.25in,clip,keepaspectratio]{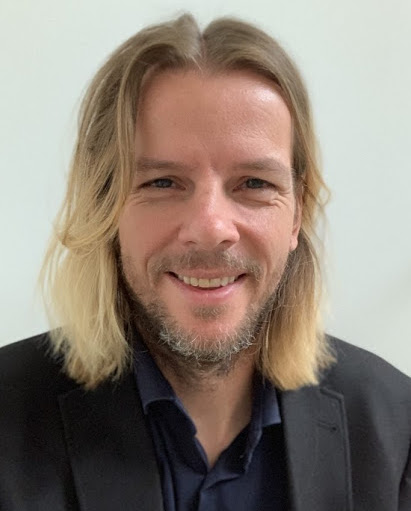}}]{Bj\"orn W. Schuller}{\,}is currently a professor of Artificial Intelligence with the Department of Computing, Imperial College London, UK, where he heads the Group on Language, Audio, \& Music (GLAM). He is also a full professor and the head of the Chair of Embedded Intelligence for Health Care and Wellbeing with the University of Augsburg, Germany, and the Founding CEO and current Chief Scientific Officer of audEERING.  He is an IEEE Fellow alongside other Fellowships. 
Contact him at \href{mailto:schuller@IEEE.org}{schuller@IEEE.org}.
\end{IEEEbiography}

\end{document}